# Reconstruction Student with Attention for Student-Teacher Pyramid Matching

Shinji Yamada[1], Kazuhiro Hotta [1]

[1] *Meijo University, 1-501 Shiogamaguchi, Tempaku-ku, Nagoya 468-8502, Japan*

Abstract: Anomaly detection and localization are important problems in computer vision. Recently, Convolutional Neural Network (CNN) has been used for visual inspection. In particular, the scarcity of anomalous samples increases the difficulty of this task, and unsupervised leaning based methods are attracting attention. We focus on Student-Teacher Feature Pyramid Matching (STPM) which can be trained from only normal images with small number of epochs. Here we proposed a powerful method which compensates for the shortcomings of STPM. Proposed method consists of two students and two teachers that a pair of student-teacher network is the same as STPM. The other student-teacher network has a role to reconstruct the features of normal products. By reconstructing the features of normal products from an abnormal image, it is possible to detect abnormalities with higher accuracy by taking the difference between them. The new student-teacher network uses attention modules and different teacher network from the original STPM. Attention mechanism acts to successfully reconstruct the normal regions in an input image. Different teacher network prevents looking at the same regions as the original STPM. Six anomaly maps obtained from the two student-teacher networks are used to calculate the final anomaly map. Student-teacher network for reconstructing features improved AUC scores for pixel level and image level in comparison with the original STPM.

## 1 INTRODUCTION

Anomaly detection is the identification of sample that deviate from the regular pattern in the data. In the industries, anomaly detection is very important technology because it can lead to increase economic sfficency. However, the scaricity of anomalous samples makes it difficult to train the model. Therefore, weakly supervised or unsupervised learning method is required for anomaly detection.

Convolutional Neural Network (CNN) [1] and Residual network (ResNet) [2] have been applied to anomaly detection [3, 4, 5]. Since those methods just classify normal samples and anomal samples, we can not obatain suitable accuracy in pixel level anomaly detection. GAN [6] and VAE [7] can detect anomalous regions at the pixel level using CNN because those methods generate normal images from the input image, and visualize the difference between the input image and the generated image. Those methods can be trained without anomalous images. Therefore, GAN and VAE have been widely used in anomaly detection. Various GAN-based [8, 9, 10, 11] and VAE-based [12, 13, 14] methods have been proposed. However, those methods require the reconstruction of image with the same resolution as the input image. If normal image is not reconstructed well, it is difficult to detect anomalous regions.

In recent years, STPM [17] which uses student-teacher network has been proposed. STPM used two networks. One is the teacher network which is the pretrained ResNet18. The other is the student network which is the untrained Resnet18. STPM detects anomalous regions by using anomaly maps at three different resolutions in the student Resnet18. However, STPM has a drawback that low accuracy in detecting cracks and other abnormalities. STPM visualizes the difference between a student network which knows only the normal and a teacher network pretrained with ImageNet [18] which can represent a variety of features. Although three anomaly maps with different resolutions are used for anomaly detection, the accuracy of anomaly maps in shallow layer with high-resolution is lower than the others. The difference in features between the teacher network and the student network is more likely to appear in the deeper layers that contain more semantic information. Since the same input image is fed into teacher and student networks, the feature maps in shallow layers are similar each other. Thus, it is difficult to differentiate between them. As a result, it is difficult to detect anomalies in high-resolution anomaly maps. The shallower layers of CNNs contain a lot of information about contours and edges. For abnormal products such as cracks, thin scratches and etc, it is easy to extract abnormal region such as edges and contours in shallower layers. Therefore, the difference between feature map in teacher and student networks become near zero, and we cannot detect the anomalies well.

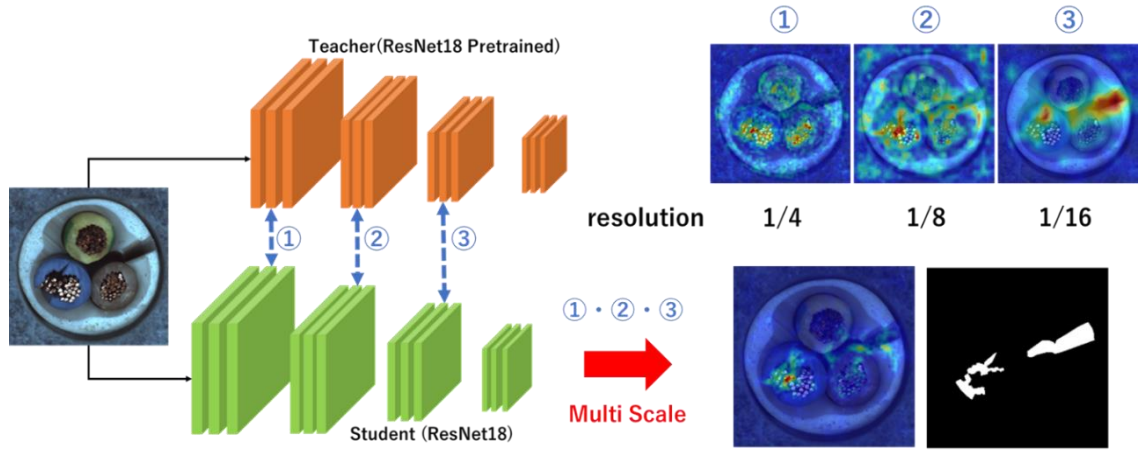

Figure 2: Overview of Student-Teacher Pyramid Matching

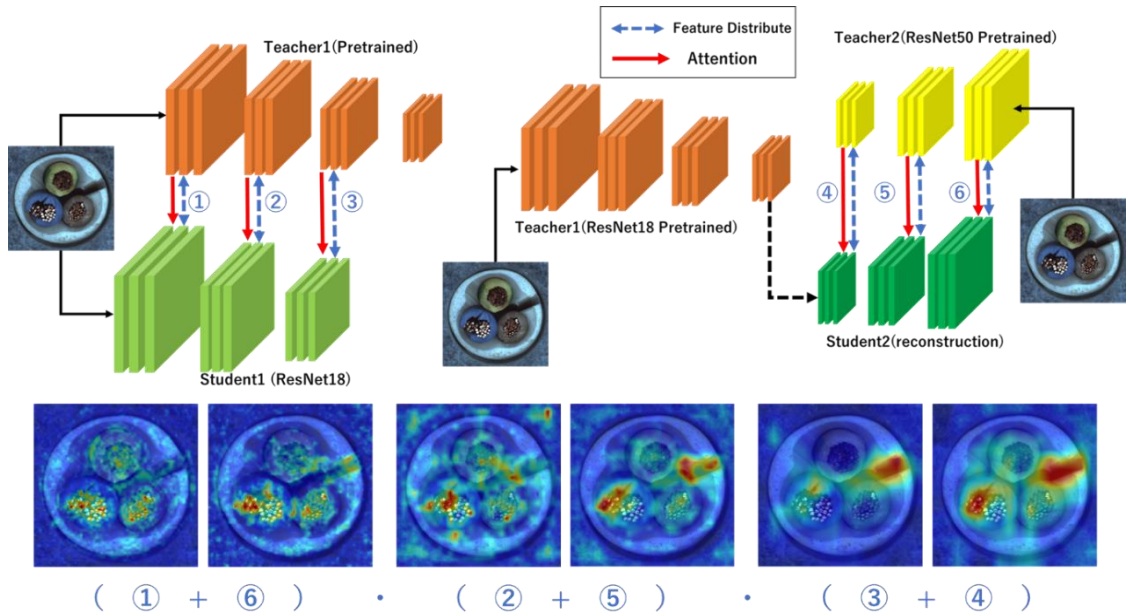

Figure 1: Overview of the proposed method

We consider that this could be prevented by using the component that reconstructs a normal image from an abnormal image. By reconstructing the features of normal products, we could make a large difference between teacher and student networks for anomalous images. In the original STPM, there were some anomalous products that could not be differentiated between student and teacher networks. The reconstruction network generates the features of normal products from abnormal product images. Then, we compute the difference between the teacher network that represents the features of the abnormal image and the student network that represents the features of the pseudo-normal product. This makes easy to detect anomalous regions in comparison with the original STPM. We then used the pretrained

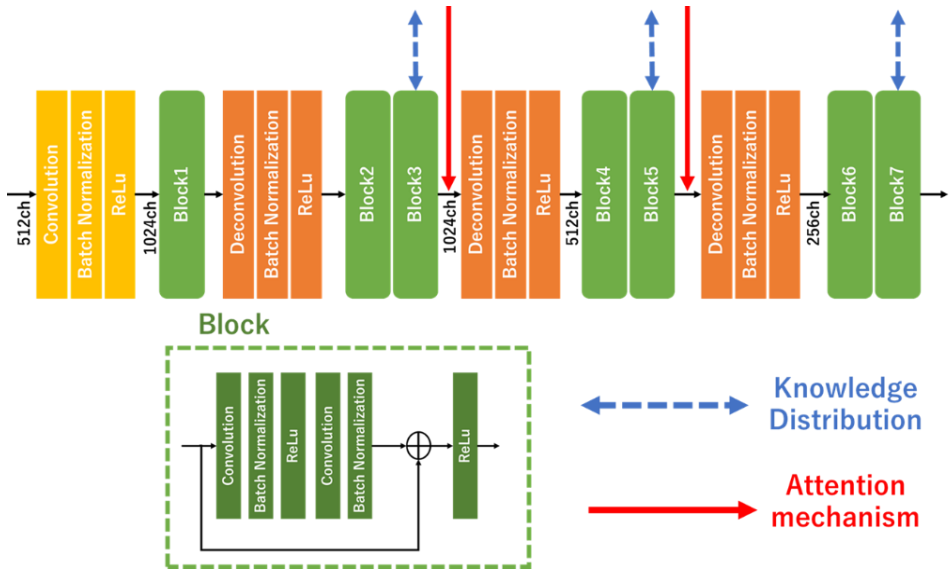

Figure 3: Structure of a student2 network

ResNet50 as the teacher network for the new student network. The original STPM used ResNet18 as the teacher network. We aim to provide a perspective that is not present in the original STPM by distilling the knowledge of the features in the ResNet50 which has a deeper structure than Resnet18. However, learning to distil the features in the ResNet50 did not work as intended. Therefore, we introduced attention mechanism between new student and teacher to improve the accuracy. Since only normal product images are used for training, the attention mechanism learns to reconstruct the normal region. In addition, attention mechanism behaves differently for abnormal regions in comparison with normal products because it has never seen abnormal regions in learning. Thus, the difference between normal and abnormal areas is emphasized by attention mechanism. This leads to improve the accuracy further in detecting abnormalities.

We evaluated our method on the MVTec Anomaly Detection datasets [29]. The pixel-level AUC of the proposed method was compared with the original STPM and outperformed it in many categories. The image level AUC of the proposed method also outperformed that of the original STPM. In addition, we show the effectiveness of different teacher network and attention mechanisms in ablation studies.

This paper is structured as follows. Section 2 describes the related works. Section 3 describes the details of our proposed method. In section 4, we show experimental results. Finally, we conclude this paper in section 5.

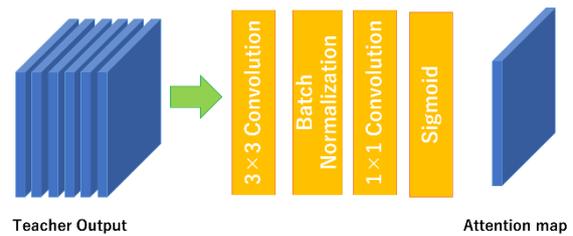

Figure :4 Details of attention mechanism

## 2 RELATED WORKS

This section describes related works. Section 2.1 describes related works of anomaly detection. Section 2.2 describes attention mechanism.

### 2.1 Anomaly detection

Anomaly detection methods are classified into two types. One is image-level anomaly detection which classifies normal and anomalous products. The other is pixel-level anomaly detection. We first mention the image level anomaly detection. There are also three groups in image-level anomaly detection. They are reconstruction-based, distribution-based, and classification-based methods. Reconstruction-based method is used for anomaly detection by reconstructing normal image. Conditional GAN [34] and VAE [31] are mainly used. Distribution-based method detects anomaly by modelling the probability distribution of normal data. If the probability distribution of an input image differs from that of the normal product images the image is detected as abnormal. ADGAN [32] is one of them. ADGAN makes it easier to estimate the probability distribution by learning a deep projection of the normal features into the latent space. Classifier-based methods include both deep learning and shallow models. Anomaly detection is performed by feeding the features of the pre-trained deep layer model into the shallow model such as one-class support vector machine (OC-SVM) [33]. However, in the inspection of industrial products, it is more important to identify which part in the product is defective than to detect abnormalities at the image level described above.

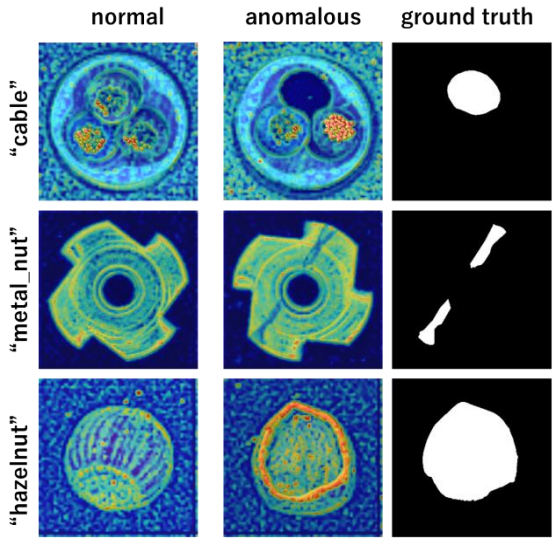

Figure 5: Examples of attention maps

Table 1: Pixel level AUC

| | Category | STPM | Ours |
|---|---|---|---|
| Textures | Carpet | 0.988 | **0.990** |
| | Grid | 0.990 | **0.993** |
| | Leather | **0.993** | 0.990 |
| | Tile | **0.974** | 0.968 |
| | Wood | **0.972** | 0.964 |
| Object | Bottle | 0.988 | **0.989** |
| | Cable | 0.955 | **0.976** |
| | Capsule | 0.983 | **0.989** |
| | Hazelnut | 0.985 | **0.991** |
| | Metal nut | 0.976 | **0.986** |
| | Pill | **0.978** | 0.971 |
| | Screw | 0.983 | **0.994** |
| | Toothbrush | 0.989 | **0.990** |
| | Transistor | 0.825 | **0.881** |
| | Zipper | **0.985** | **0.985** |
| | Mean | 0.970 | **0.977** |

Table 2: Image level AUC

| STPM | Ours |
|---|---|
| 0.955 | **0.969** |

Pixel-level anomaly detection is effective for the visual inspection. Anomaly detection at the pixel level is more complex than at the image level, because visual anomalies should be captured at the pixel level. GAN-based [8, 9, 10, 11] and VAE-based methods [12, 13, 14] are often used for pixel-level anomaly detection methods. Those methods detect anomalous regions by visualizing the difference between the input image and the reconstructed image. However, GAN-based methods have the possibility of vanishing gradients and VAE has the problem of blurring the generated image even if the loss is optimized. Moreover, if the GAN or VAE cannot reconstruct normal image well, anomaly detection does not work well.

Different methods from GAN and VAE have also been studied. In recent years, several methods using the pretrained model have been proposed as new approches. SPADE [15] is one of them. Multiple normal images are fed into the pretrained model, and features are obtained. The feature map of anomalous image are also obtained from the pretrained model, and we compute the difference between the feature map of the anomalous image and each feature map for multiple normal images. Finally, the difference maps are averaged to obtatain an anomaly map. Since the SPADE uses multiple normal images at test phase, computational cost is high. On the other hand, Student-Teacher framework [30] has also proposed for anomaly detection. Uninformed students [16] is a method that uses the student-teacher framework to obtain an anomaly map. Anomaly maps are calculated from two perspectives. One is the difference between the outputs of student and teacher networks. The other used the uncertainty of the output of multiple student networks. However, multiple student networks are trained separately in this method. This makes learning inefficient.

Table 3: AUC score at each resolution

|  |  | 1/4 | 1/8 | 1/16 | Multi Scale |
|---|---|---|---|---|---|
| Pixel level AUC | STPM | 0.915 | 0.953 | 0.957 | 0.970 |
|  | Ours | **0.935** | **0.966** | **0.967** | **0.977** |
| Image level AUC | STPM | 0.808 | 0.917 | 0.934 | 0.955 |
|  | Ours | **0.858** | **0.951** | **0.953** | **0.969** |

STPM is a method that uses teacher-student networks; Teacher network is the pretrained model with ImageNet and only the student network needs to be trained. Since teacher network can represent anomalous regions well and student network cannot represent them, the difference in features of two networks for anomalous regions becomes large and we can detect anomalous regions. The processing speed of STPM is higher than the SPADE using many normal samples. However, as mentioned in section 1, STPM also has its drawbacks. STPM used the difference between the features of the student network and the teacher network as the anomaly score, but anomaly score for anomalous products such as cracks becomes small because student network can repersent edges and contours well.

We introduce a new student network that reconstructs the features of normal products to address the problem. The new student network tries to reconstruct the features of normal products from an abnormal image. Therefore, we emphasize the difference between the teacher network that represents the features of abnormal images and the network that represents pseudo-normal features. This make our method easy to detect anomalous regions.

## 2.2 Attention mechanism

In image recognition, various attention mechanisms have also been proposed. Residual Attention Network [19] solves the vanishing gradient problem by using a similar structure to a residual block. Squeeze-and-Excitation (SENet) [20] introduced an attention mechanism which emphasizes important channels in feature maps. As a variant of SENet, accuracy booster blocks [21] and efficient channel attention module [22] were proposed. Those methods changed fully-connect-layer in SENet. Transformer [23] was proposed for language translation using only attention mechanism. Self-attention is a kind of transformer, and several image recognition methods using self-attention have been proposed [24-26]. Attention branch network [28] proposed an attention map for classification by aggregating multiple feature maps. The attention map can be used to visualize the basis for decisions. By using classifier branch before attention map, attention map had a basis for classification decision based on Class Activation Mapping (CAM) [27].

Table 4: Experimental results with different teacher networks

|  | STPM | ResNet18 | ResNet50 | ours |
|---|---|---|---|---|
| Pixel level AUC | 0.970 | 0.973 | 0.976 | **0.977** |
| Image level AUC | 0.955 | **0.972** | 0.961 | 0.969 |

Table 5: Accuracy of the proposed method w/o attention mechanism

|  | STPM | Without attention | Ours (With attention) |
|---|---|---|---|
| Pixel level AUC | 0.970 | 0.973 | **0.977** |
| Image level AUC | 0.955 | 0.954 | **0.969** |

In this paper, the purpose of attention mechanism in our method is to leak the features of teacher network to the student network in order to reconstruct the features of normal products well. Since the anomaly map must detect anomalies at the pixel level, attention mechanism that can emphasize and suppress pixels is more suitable than an attention mechanism that emphasizes channels as in SENet. Attention mechanism in our method is used to leak some of the features of the teacher network which is the correct answer to the students. If we leak the most of features in teacher network to student network, there will be no difference between student and teacher networks. Therefore, by aggregating the features in the teacher network into one channel, it is possible to emphasize and suppress the pixels without giving all information. Thus, we use an attention map generated from teacher network. In this paper, since only normal products are used for learning, the attention mechanism is used for reconstructing the normal region.

# 3 PROPOSED METHOD

This section describes the details of the proposed method. Our proposed method has three contributions; Student network for reconstruction, attention from teacher network to student network and different teacher network from original STPM.

Section 3.1 describes the original STPM and its problem. Section 3.2 describes student network for reconstruction and its teacher network. Section 3.3 describes attention mechanism in the student network for reconstruction.

## 3.1 STPM and its problem

Figure 1 shows an overview of the original STPM. STPM used two networks for anomaly detection. Teacher network is the ResNet18 pretrained with ImageNet. The student network is the untrained ResNet18. Knowledge distillation is performed at the feature maps with three different resolutions, and the student network is trained to resemble the features in the teacher network. At test phase, anomaly maps are obtained by taking the difference between feature maps of teacher and student networks at three different resolutions. Figure 1 shows anomaly maps at three different resolutions. Three anomaly maps are multiplied to obtain a final anomaly map.

We explain the loss function of original STPM because we use the same loss functions. The normal image is defined as $I_k \in \mathbb{R}^{w \times h \times c}$ where h is the height, w is the width, c is number of channels. We normalize the features in teacher and student networks along the channel dimensions as

$$\hat{F}_t^l(I_k)_{ij} = \frac{F_t^l(I_k)_{ij}}{\left\|F_t^l(I_k)_{ij}\right\|_{l_2}} \quad (1)$$

$$\hat{F}_s^l(I_k)_{ij} = \frac{F_s^l(I_k)_{ij}}{\left\|F_s^l(I_k)_{ij}\right\|_{l_2}} \quad (2)$$

where $F_t^l$ and $F_s^l$ represent the features in teacher and student networks. $l$ represents the resolution of feature map and $(i,j)$ is position. The student network learns to have the same normalized output. The loss function is defined as

$$L^l(I_k)_{ij} = \frac{1}{2}\left\|\hat{F}_t^l(I_k)_{ij} - \hat{F}_s^l(I_k)_{ij}\right\|. \quad (3)$$

Since the same loss function is used for feature maps at three different resolutions, the final loss function is as follows.

$$L^l(I_k) = \frac{1}{w_l h_l}\sum_{i=1}^{h_l}\sum_{j=1}^{w_l} L^l(I_k)_{ij} \quad (4)$$

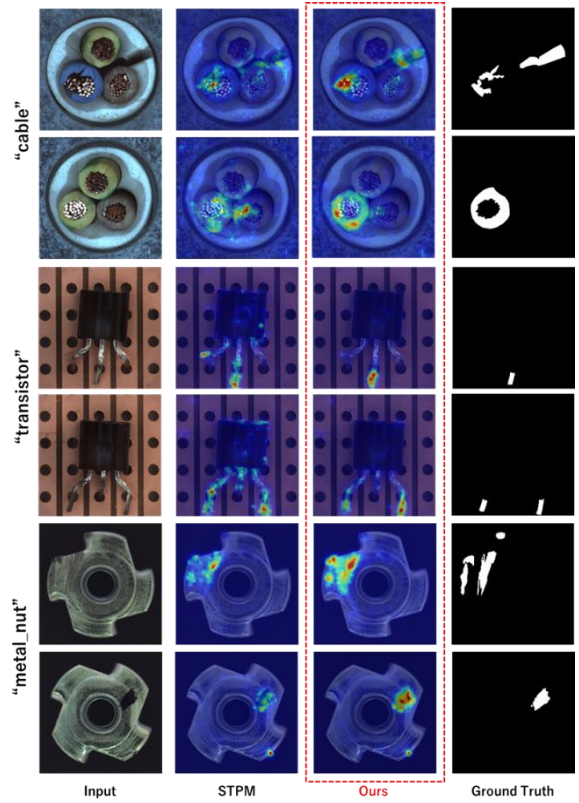

Figure 6: Results of anomaly mapping

STPM used three different resolutions; 1/4, 1/8, 1/16 of the input image.

In test phase, anomaly maps are calculated at three feature maps with different resolutions. Three anomaly maps are resized to the same size as the input image, and the final anomaly map is obtained by multiplying the three maps.

However, STPM has drawbacks as shown in Figure 1. The accuracy of the anomaly map with the higher resolution is poor. Figure 1 shows the fact. That the anomaly maps 1 and 2 in the Figure do not detect any anomalies. STPM computes the difference between the teacher network which has rich feature representation by pretraining with ImageNet and the student network which only knows normal products. These are more likely to appear in the deeper layers that contain semantic information. Conversely, since the same input image is fed into both student and teacher networks, the features in the shallow layers closer to the input image are similar. As a result, the anomaly detection accuracy at the feature maps in shallow layer with high resolution becomes poor though the multiplication of three anomaly maps was the better accuracy than each anomaly map in [17]. As mentioned in section 1, such problem is more likely to occur with abnormal products such as scratches on the line or cracks shown as examples in Figure 1. Since CNN extracts edges and contours of objects in shallow layers, the features of abnormal products such as scratches and cracks tend to be extracted in shallow layers. As a result, the difference between student and teacher networks is small for such anomalous products. The usage of only the anomaly map at deep layer does not improve the anomaly detection accuracy. Some anomalies may not be detected or may be over-detected only by the anomaly map at deep layer. Therefore, it is necessary to improve the anomaly map in shallow layers.

It would be better if the student networks were less expressive of anomalous products. By reconstructing the feature maps of normal products in new student network, we can emphasize the difference between the teacher network that represents the anomalous parts and the student network that reproduces the features of pseudo normal products. We consider that this would ultimately lead to improve the accuracy of anomaly detection.

### 3.2 Proposed student network

We propose a visual inspection method using new student network that reconstructs features. There are three main contributions in the proposed method. The first one is the addition of the reconstruct network. The second one is that we use a different teacher network from the original STPM for new student network. The third one is the attention explained in next section.

Figure 2 shows the overview of the proposed method. The input of new student network is the feature map at lowest resolution in the pretrained ResNet18, and student network reconstructs features of normal products. The computational graph is disconnected between the teacher1 and student2, and only the student2 is trained by knowledge distillation.

The structure of new student network is the similar as the Resnet18 which has residual structure. Therefore, if we use the same teacher network for student2 as the original STPM, it may make the same mistakes as original STPM. Thus, we use ResNet50 pre-trained with ImageNet as the teacher network for student2 because ResNet50 has different features and structure from ResNet18 which is the teacher1. We expect to have a different perspective from the conventional method, and such features will improve the anomaly detection accuracy.

In the proposed method, the student1 in the original STPM with attention mechanism and the student2 for reconstruction are trained separately. Equations 1 to 4 as shown in Section 3.1 are used for knowledge distillation. The input of student2 is the feature maps with a size of 1/32 of the input image, and student2 reconstructs the feature maps up to 1/4 of the size of input image. The anomaly maps 1-3 in Figure 2 are obtained from the original STPM with attention mechanism. The anomaly maps 4-6 obtained from the reconstruction network are also shown. We see that the anomaly map 4 is the lowest resolution. When we compare anomaly maps of the original STPM with those of the reconstructed network, we see that the accuracy of the reconstructed network is better than that of the original STPM. In particular, when we compare the anomaly maps 1 with 6 which have the highest resolution, anomaly map 6 by the reconstructed network was able to detect anomalies that were not detected by the original STPM.

In test phase, six anomaly maps are obtained by student1 and 2. As shown in Figure 2, the anomaly maps of the same resolution are added together, and then the anomaly maps of different resolutions are multiplied and final anomaly map is generated. The anomaly detection is performed using the final anomaly map.

### 3.3 Attention mechanism for student network

The third contribution in the proposed method is the attention mechanism between new student network and teacher network. This mechanism acts so that the features in the student network in the normal region are close to the features in the teacher network. Figure 4 shows the structure of the attention mechanism. The attention mechanism receives the feature maps of the teacher network and aggregates the feature maps into one channel. If we leak the most of features in teacher network to student network, there will be no difference between student and teacher networks. Thus, we used an attention map with only one channel to limit the information given. We consider that the attention mechanism which emphasizes the channel would not be very effective in detecting anomalies because it could not learn important locations for pixel-level anomaly detection. We consider that one attention map would be better because it could emphasize the important pixels. The computational graph between the attention mechanism and the teacher network is not connected, and teacher network is not updated. Since only normal products are used as training data, the attention mechanism learns to successfully restore the normal region.

Figure 5 shows the attention map when anomalous products are used as inputs. We show the attention maps at the highest resolution which are feature map 6 in the Figure 2. Figure shows the three categories of normal and abnormal products, and the ground truth labels corresponding to the abnormal products. By comparing the attention maps of normal products with those of abnormal products, the attention maps of abnormal products in "cable" and "metal_nut" did not respond to the abnormal area, but respond to the normal area. In "hazelnut", the attention map reacts to the abnormal area. Although the behaviour differs depending on the category, attention response is different between normal and abnormal areas. The network is trained with only normal products. Therefore, it behaves differently from the normal product when the input is abnormal data that has never been seen during training. We use the attention mechanism for both conventional STPM and reconstruct networks. We introduce the attention mechanism into all resolution blocks where knowledge distillation is performed. The usage of attention mechanism encourages that student network represents the features in the normal regions well.

## 4 EXPERIMENTAL RESULTS

In this section, we show the experimental results. Section 4.1 describes datasets and experimental conditions. Section 4.2 shows experimental results on the MVTech dataset. Section 4.3 shows the results of ablation study.

### 4.1 Datasets

In this paper, we use the MVtec anomaly datasets [29] for experiment. MVtecAD includes 15 categories of industrial products. The training data are only normal products. The image size is different for each category, and we resize the images to 256 x 256 pixels. We train conventional STPM with attention mechanism and our proposed method for 100 epochs. We use the SGD optimizer with the momentum of 0.9. We set the learning rate to 0.4, the batch size to 32, and the weight decay to $10^{-4}$. We use the pixel-level and image-level AUC (Area under the ROC curve) as evaluation measures.

### 4.2 Results on Anomaly detection

First, we describe the pixel-level AUC score. Pixel-level AUC is an indicator that shows whether the anomaly map is correctly detecting the anomalous area or not. Table 1 shows the results of pixel-level AUC. We see that our method outperforms conventional STPM in many categories. In particular, it significantly outperforms the conventional method in the categories of "cable" and "transistor". Cable includes anomalies such as cracks as described in Section 3.1. The network that reconstructs the features worked effectively led to the improvement of the anomaly detection accuracy. Although we show anomaly maps later, the anomaly maps of the proposed method have less noise than conventional STPM. The new student network makes the anomaly map more accurate.

Next, we show the results of the image-level AUC scores. Image-level AUC is a score that indicates whether the input image can be correctly classified as an abnormal product or not. Table 2 shows the image-level AUC. The accuracy of the proposed method is higher than that of original STPM. Anomaly detection at image level is performed using the maximum value in an anomaly map. To compute AUC, threshold for classifying normal and abnormal is changed. Since the anomaly map of the proposed method is more accurate than conventional STPM as shown in Table 1 and Figure 6, the accuracy of anomaly detection at the image level was also improved.

We also evaluated the accuracy of anomaly detection at each resolution. Table 3 shows pixel-level AUC and image-level AUC at each resolution. We see that AUC score of the proposed method exceeds the original STPM at all resolutions. In particular, the pixel-level AUC and image-level AUC at the top two resolutions are much improved. Conventional STPM had poor anomaly detection accuracy at high resolution. As a result, anomalous regions were missed in the final anomaly map, which was created by multiplying the three anomaly maps. As shown in Table 3, the proposed method improved the accuracy of anomaly detection for all three anomaly maps, especially for high resolution. The accuracy improvement is derived from attention mechanism and reconstruction.

Finally, we show anomaly maps of conventional STPM and the proposed method in Figure 6. The proposed method is able to detect anomalies such as cracks which are the drawbacks of conventional STPM. In addition, the proposed method is able to accurately identify the anomalies in the cables of different colors while original STPM is not able to detect them. For transistor category, conventional STPM produces an anomaly map for the whole transistor's legs. In contrast, the anomaly map in our method detects only the anomalous part of the transistor's legs. When we compare the anomaly maps with ground truth, our method is obviously better than the conventional method. Finally, the results for the metal nut are shown. For anomalous products of scratches and discoloration on a metal nut, STPM detected only a part of the anomalies. In contrast, the proposed method could accurately detect anomalies. These results demonstrated that our proposed method improves conventional STPM.

### 4.3 Ablation study

In the proposed method, the student network 1 and the student network 2 used different teacher networks as shown in Figure 2. We also used attention mechanism from the teacher network to the student network. First, we discuss the impact of changing the teacher network. Table 4 shows the results when we use the same teacher network (ResNeT18, ResNeT50) in student network 1 and 2. By comparing the results, we see that our method with the different teacher networks has better accuracy at both pixel levels. The detection accuracy at the image level is higher when ResNeT18 is used as the teacher network for both networks. However, the pixel level AUC is not high. In contrast, the proposed method has high accuracy in detecting anomalies at the pixel and image levels. Therefore, we consider that our method using a different teacher network is better. Different teacher networks lead to a different perspective from the conventional method, and anomaly detection accuracy was improved.

Next, we evaluate our method with/without the attention mechanism. Table 5 shows that our method using the attention mechanism has better anomaly detection accuracy than our method without attention mechanism. Attention mechanism is useful to reconstruct normal areas. This enhanced the difference between anomalous and normal areas. When there is no attention mechanism, it cannot make the difference and the accuracy is inferior to the proposed method.

## 5 CONCLUSIONS

In this paper, we improved the anomaly detection method STPM by using new student network for reconstruction. We also used attention mechanism from the teacher network to the student network. When an anomalous product image is fed into the proposed method, the reconstruction network works to convert the features of the abnormal products into the features of normal products. Thus, the student network will output pseudo-normal product features, and this makes it easier to differentiate between the abnormal product features obtained from the teacher network. In addition, we used the ResNeT50 as a teacher network for the reconstruction network. We can give the student network to a different perspective from the conventional method. In addition, we used the attention mechanism to help the student network represent the features of normal regions. This attention mechanism showed the difference between normal and abnormal areas, which led to improved accuracy in detecting abnormalities. The effectiveness of our proposed method is demonstrated by the comparison with the original STPM. We also showed the effectiveness of attention mechanism and different teacher network by the ablation studies.

Here we introduced a new student network for reconstruction and the anomaly maps obtained from the network are combined with anomaly maps in conventional STPM with attention mechanism. We used simple summation of anomaly maps of student1 and 2. There may be better integration method. This is a subject for future works.